\documentclass{article}

\usepackage{iclr2020_conference,times}
\iclrfinalcopy

\usepackage[utf8]{inputenc} 
\usepackage[T1]{fontenc}    
\usepackage{hyperref}       
\usepackage{url}            
\usepackage{booktabs}       
\usepackage{amsfonts}       
\usepackage{nicefrac}       
\usepackage{microtype}      

\usepackage{csquotes}

\usepackage{amsmath}	    
\usepackage{graphicx}
\usepackage{amsthm}
\usepackage{mathtools}

\usepackage{enumitem}

\title{Towards intervention-centric causal reasoning in learning agents}
\author{Benjamin Lansdell\\Department of Bioengineering\\University of Pennsylvania\\
Philadelphia, PA}
\begin{document}

\maketitle

\begin{abstract}
Interventions are central to causal learning and reasoning. Yet ultimately an intervention is an abstraction: an agent embedded in a physical environment (perhaps modeled as a Markov decision process) does not typically come equipped with the notion of an intervention -- its action space is typically ego-centric, without actions of the form `intervene on X'. Such a correspondence between ego-centric actions and interventions would be challenging to hard-code. It would instead be better if an agent learnt which sequence of actions allow it to make targeted manipulations of the environment, and learnt corresponding representations that permitted learning from observation. Here we show how a meta-learning approach can be used to perform causal learning in this challenging setting, where the action-space is not a set of interventions and the observation space is a high-dimensional space with a latent causal structure. A meta-reinforcement learning algorithm is used to learn relationships that transfer on observational causal learning tasks. This work shows how advances in deep reinforcement learning and meta-learning can provide intervention-centric causal learning in high-dimensional environments with a latent causal structure.
\end{abstract}

\section{Introduction}

\begin{displayquote}"...suppose that an individual ape ... for the first time observes the wind blowing a tree such that the fruit falls to the ground... we believe that most primatologists would be astounded to see the ape, just on the basis of having observed the wind make fruit fall ... create the same movement of the limb ... the problem is that the wind is completely independent of the observing individual and so causal analysis would have to proceed without references to the organism’s own behavior." Tomasello and Call, 1997 \citep{tomasello1997primate}
\end{displayquote}

Learning causal relationships in an environment is necessary for flexible planning and problem solving. Humans are adept at learning the causal structure of an environment, not just from their own actions, but also through passive observation \citep{Woodward2007a}. This includes imitating or taking cues from other animals (social learning), but also learning causal relationships in environments where no other agents are present. For instance, when a broom resting against a wall accidentally falls and hits a light switch, and a light subsequently turns on, we may infer that the switch causes the light to turn on. It has been argued that this ability is related to an intervention-centric view of causality, which makes us significantly more flexible observational learners and causal reasoners than other animals \citep{Woodward2007a,Woodward2010}.

According to the interventionist account of causality, a common view in statistics, philosophy and psychology \citep{gopnik2007causal}, a causal relationship exists between two events if intervening to make one occur results in the other occurring \citep{Woodward2003}. Importantly, an intervention is an abstract notion -- it does not matter what is intervening, just as long as it is somehow external to the system being studied. In the light switch example, the accidentally-falling broom may be viewed as an intervention on the ‘switch-light’ system. Learning how to turn on the light by observing this scene can be achieved with the following things: first, by viewing the environment as one that has causal structure, that exists independently of the agent; second, understanding that some of the agent’s actions may be interventions that can exploit this structure; and, third, understanding that other objects/agents in the environment may also be able to intervene to exploit causal relationships. With these an agent could learn, just by observing the falling broom, that if it acts in a way to intervene on the light switch then the light will turn on. Since the notion of intervention is abstract -- it could be the broom or the agent -- this allows the agent to transfer what it observes to what will happen when it acts on the world in a very flexible manner (Figure \ref{fig:overview}a).

How else might causal relationships be learnt and exploited? Consider other ways an animal (or a more generic agent) might learn how to turn the light on. First, it could learn directly from its own experience, by turning the switch on itself. Any agent that learns through operant conditioning may exploit causal relationships in this manner \citep{Gershman2015}. And, second, it could learn by imitating or otherwise being directed to the switch by another animal (or agent). In fact these are the dominant forms of learning in the learning agent literature -- reinforcement learning and imitation learning, respectively \citep{Sutton2017,Edwards2018}. While they may result in an agent learning to use the causal structure of the environment, they do not require the agent possess a sense of interventions. Thus such agents cannot learn from observation in the same way as the intervention-centric learner described above. 

\subsection{Intervention-based causal learning}

What abilities are indicative of an intervention-centric notion of causal relationships? One telling ability is the ability to integrate information from observation and an agent's own actions into a single causal model of the environment, and to use this to execute novel actions in order to obtain some desired outcome. By being novel, the action could not have been learnt through operant conditioning, nor could it have been suggested by watching another agent. Instead, it must have come from predicting a desirable outcome based on an understanding that some observed relationships will also hold when the agent performs certain actions, that is, it must have come from predictions based on a causal model of the environment. In the animal kingdom, humans are superior tool users and observational causal learners. Corvids for instance, despite their cunning, seemingly are not able to create novel interventions from observation \citep{Taylor2014,Taylor2015}. The notion of an intervention thus appears as a powerful and defining characteristic of human intelligence -- it underlies our ability to understand and manipulate the world \citep{Woodward2003}. Yet tests of such abilities in the learning agent literature are largely lacking; the focus is instead typically on reinforcement learning or imitation learning. It remains relatively unexplored how this form of causal learning can be implemented in a learning agent setting.

Of course, one way is through an explicit causal model. For learning agents whose environment model is a causal Bayesian network (CBN) \citep{pearl2000causality}, or structural causal model (SCM) \citep{Peters2017}, an intervention-centric sense of causality is hard-coded into the structure of the model and action space. That is, the environment factors into a network of cause-effect relationships, and the action space may straightforwardly be given as interventions on edges of that network. In these cases, observational learning is a very well explored problem with many different approaches \citep{pearl2000causality,Zhang2016a,Peters2017,Bareinboim2015,Bareinboim2016,Zhang2017a}. The problem is that SCMs are hard to scale to high-dimensional state spaces, perhaps with a latent causal structure, and thus they are hard to combine with modern deep-learning-based methods (though see \cite{Arjovsky} for interesting ideas). Thus we may want to look for other ways of constructing learning agents with intervention-centric learning abilities. 

\subsection{Contribution}

\begin{figure}[t]
\centering
\includegraphics[width=0.9\textwidth]{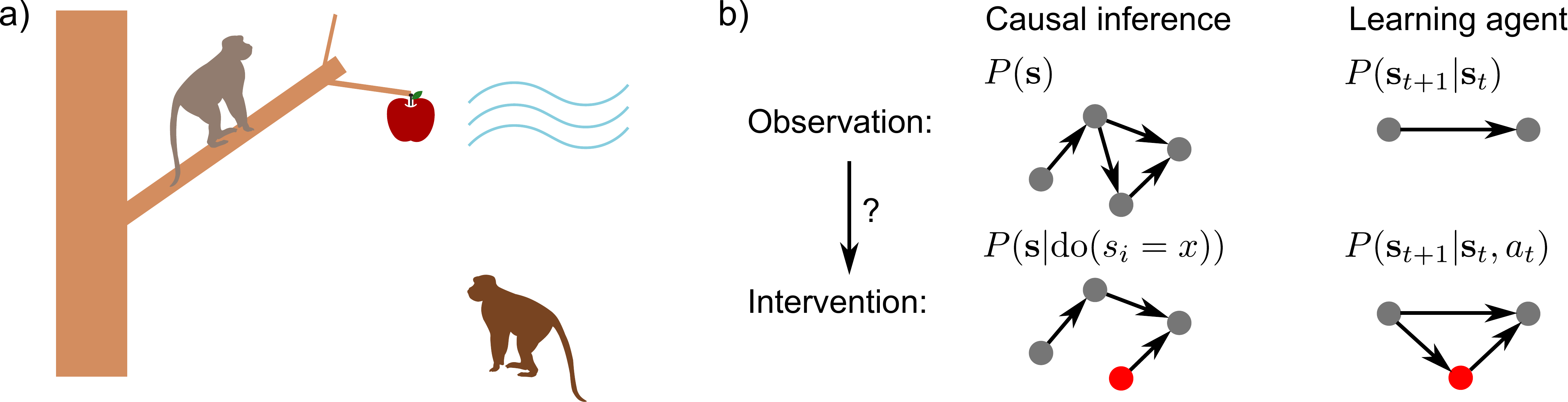}
\caption{a) A causal relationship exists between the state of the branch and the state of the apple: if the branch is shaken, the apple can fall. By definition, it does not matter what intervenes on the branch to make it fall. This permits learning the causal relationship from a diverse range of sources: self action, another's action, or some other perturbation. b) Causal inference assumes an observational distribution, and asks when and how can we learn about what happens when intervening on the environment. Here, we argue the analogous setting in learning agents is to assume access to an observational distribution, and to study when and how this can be transferred to learn policies over action-conditioned distributions.}
\label{fig:overview}
\end{figure}

We focus on this problem in a reinforcement learning agent setting, i.e. with Markov decision processes (MDPs). We will break down our analysis into two Problems:
\begin{enumerate}
\item How can an agent learn a causal model from observation? 
\item How can an agent learn which sequence of its actions constitute an intervention that can exploit the causal model it has learnt? 
\end{enumerate}
Here learning from observation means the agent observes only state transitions, $\mathbf{o}_t$, and not the actions that may have been taken by some other entity to generate them. In what sense does this get at intervention-centric causal learning? By transferring from an observational setting to a setting where the agent is able to act on the environment, the agent exploits the fact that the observational and action phases share a common causal structure. This can be thought of as analogous to the typical problem in causal inference: how to transfer information from an observational distribution to candidate interventional ones (Figure \ref{fig:overview}b). While Problem 1 has been studied \citep{Nair2019,Thomas2017,Sawada2018}, to our knowledge Problem 2 has seen less attention. Yet arguably solutions to both are needed for a flexible learner that can fully exploit what it learns from observation and action. 

We use a meta-reinforcement learning approach. The aim is that the agent learns for itself both what patterns of covariation indicate transferable relationships within the environment, and which of its actions may be able to exploit these relationships. In this way the agent performs causal learning in a flexible manner, without explicitly casting its environment into causal factors, and without having explicit labels that indicate when data is being drawn from different interventional distributions.

\section{A motivating example}

We start with a simple example. Consider a state space that factors into three variables: $\mathbf{s}_t = (s_t^1, s_t^2, s_t^3)$. Then suppose the environment has one of the two dynamic causal structures, where the differences between Model A and B is highlighted in \textbf{bold}:
\begin{itemize}
    \item Model A (chain):
\begin{align*}
s^1_t &= y_t^1\\
s^2_t &= y^2_t + (1-y^2_t)(s^1_{t-1})\\
s^3_t &= y^3_t + (1-y^3_t)(\mathbf{s^2_{t-1}}),
\end{align*}
where $y_t^i \sim \text{Bn}(p_i)$ denote Bernoulli random variables with probability of activation $p_i$. 
\item And Model B (delayed fork):
\begin{align*}
s^1_t &= y_t^1\\
s^2_t &= y^2_t + (1-y^2_t)(s^1_{t-1})\\
s^3_t &= y^3_t + (1-y^3_t)(\mathbf{s^1_{t-2}}),
\end{align*}
\end{itemize}

Importantly, it is possible with certain parameter choices that, even though the models have a different causal structure, if all an agent observes is $\mathbf{s}_t$, it could not distinguish between Model A and Model B. It would have to interact with the system to tell the two apart. Consider then the task of learning which is the true causal structure. We will explore this problem using the meta-reinforcement learning approach of \cite{Wang2016c}, i.e. the same approach taken in \cite{Dasgupta2018} for causal inference through meta-RL, though they test their approach on static causal environments. 

\subsection{Meta-learning agent architecture}

The meta-reinforcement learning approach works as follows. A task, or trial, is sampled from some distribution $\mathcal{D}$. Here, this just means one of Model A or B is chosen with probability $p_A = 0.5$. Within the trial, states are generated according to the appropriate model, for $N$ steps. An LSTM network \citep{Hochreiter1997} (with 48 hidden units) was used. At each time step the LSTM receives the vector ($\mathbf{s}_t,\mathbf{a}_{t-1},r_{t-1}$) as input, where $\mathbf{s}_t$ is the observation, $\mathbf{a}_{t-1}$ is the previous action (as a one-hot vector) and $r_{t-1}$ the reward (as a single real-value). The outputs are a linear function of the LSTM’s state. A set of logits are output (with dimensionality equal to the number of available actions), plus a scalar baseline. A softmax is applied to the logits, and then sampled to give a selected action. Learning was by the asynchronous advantage actor-critic (A3C) \citep{Mnih2016} algorithm. In this framework, the loss function consists of three terms: the policy gradient, the baseline cost and an entropy cost. The baseline cost was weighted by 0.05 relative to the policy gradient cost. Optimization was done by ADAM optimization with learning rate $10^{-3}$, $\beta_1 = 0.9$, $\beta_2 = 0.999$, $\epsilon = 10^{-7}$. What follows was implemented in TensorFlow.

\subsection{Task details}

The agent is presented with $N=10$ observations from a sequence $\mathbf{s}_t$ generated according to either Model A or Model B. The agent is rewarded at the end of the trial for correctly identifying the true model. That is, the action space is $\mathcal{A} = \{A,B\}$. If $a_N$ is the correct model, then reward $r_N = 1$ is given. Otherwise $r_N = 0$. At all other times in the trial, the reward is zero. We compare performance of this meta-RL approach in four different settings, each providing different amount of information to the agent, and allowing different levels of interaction with the environment. The specific values for these models are provided in the supplementary material.

\begin{enumerate}
    \item Confounded -- the agent only observes $\mathbf{s}_t$, thus has insufficient information to solve the problem. 
    \item Observational -- there are now perturbations of the environment, changing the distribution over $\mathbf{s}_t$, but these perturbations are unobserved. More specifically, $z^i_t \sim \text{Bn}(p^{int}_i)$ are `intervention indicator' variables sampled at each time step. If $z^i_t = 1$ then $y^i_t = 1$, otherwise it follows the same dynamics as before. For some cases, this may allow the agent to identify the true causal structure.
    \item Off-policy interventional -- the agent now observes the perturbations $\mathbf{z}_t$, concatenated onto $\mathbf{s}_t$. Now the correct structure is identifiable, provided the agent can learn which of $z^2$ and $z^3$ are associated with each node $s^2$ and $s^3$. 
    \item On-policy interventional -- now the action space of the agent is the perturbation-space $\mathbf{z}$. In this case, the action space for the agent can be considered as interventions on the environment. Here the agent additionally observes a go cue, at the end of the trial, to indicate it should provide its response as to which is the correct underlying model.
\end{enumerate}

\subsection{Results and discussion}
\begin{figure}[t]
\centering
\includegraphics[width=\textwidth]{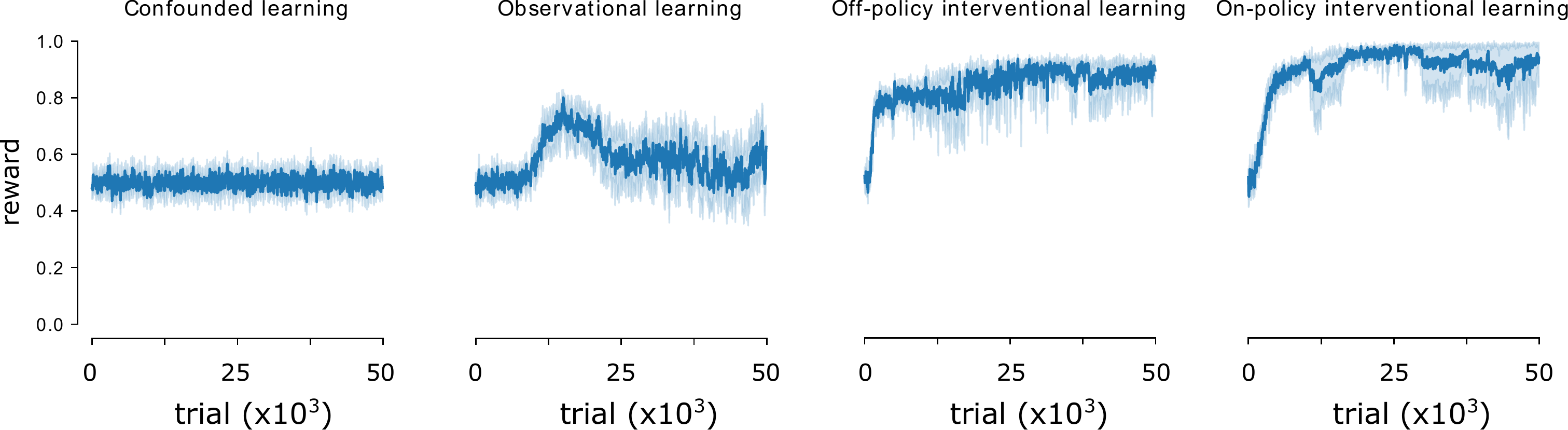}
\caption{Average reward for meta-RL agent in simple causal inference environments. Curves show mean plus/minus standard error over $n = 10$ runs.}
\label{fig:motivating_ex_results}
\end{figure}

Though this is a very simple example, it captures different forms of causal learning. In the purely observational setting, in this case, no causal learning is possible. In the observational with perturbations setting, causal learning may be possible, and is akin to something like emulative learning with `ghost conditions' in cognitive science (e.g. the wind blowing the branch thought experiment above) \citep{Hopper2010}. In the off-policy interventional setting, causal learning may again be possible, and may be aided by learning to exploit the additional observed cues that a variable is being perturbed from its default dynamics, $\mathbf{z}$. Finally, in the on-policy interventional setting, the agent can learn the causal model directly through its own interventions -- agent-centric causal learning possible through any reinforcement learning algorithm. The results using the meta-RL algorithm show both the on- and off-policy interventional settings are indeed solvable after a few thousand trials (Figure \ref{fig:motivating_ex_results}).

This setup allowed us to explore causal learning under two significant Assumptions:
\begin{enumerate}
    \item The causal relationships are expressed more or less directly between the observed variables, the causal structure is not latent.\footnote{Technically, the delay introduces some degree of unobservedness.} Further, the variables indicating whether and which variable was being perturbed were also directly observed, at least in the relevant setting. 
    \item The action space is given as direct manipulations of the underlying causal variables.
\end{enumerate}
These assumptions relate to Problems 1 and 2 described in the introduction. Thus, to extend current approaches, we consider ways in which these assumptions can be dropped. 

\section{Dealing with latent causal structure}

We can first study settings that relax Assumption 1. In fact this is very close to the problem tackled by \cite{Nair2019}, except in their case the focus is on planning to reach a goal state. And in their case the actions can be taken as direct manipulations of the underlying causal variables, an assumption we will drop in the next section.

\subsection{Task details}

To extend our simple example to this case, we consider an 8x8x3 observation space, denoted $\mathbf{o}_t$. Three pixels in this space correspond to the state of $s^1_t, s^2_t, s^3_t$, obeying the same dynamics as in the previous section. When active, the corresponding pixel is colored white, otherwise it is black. Then, to mix the state of these variables in the observations, a Gaussian blur is applied to the image. In the relevant conditions, the additional indicator variables $\mathbf{z}_t$ are added in the same locations as $s^i$, but are added only to the red channel. Further, the observed image at each frame $t$ is a 50-50 mix of the previous frame $\mathbf{o}_{t-1}$ and the current one (the Gaussian blurred pixels corresponding to the state of the system at frame $t$). This introduces some temporal blurring in the observed dynamics also (Figure \ref{fig:latentresults}a). Now the relationship between the variables $y_i$ and the observed images is less obvious. As before, reward $r_N = 1$, is given for correctly identifying the underlying model, and $r_N = 0$ otherwise. The observed image $\mathbf{o}_t$ is fed to the learning agent, instead of the underlying state $\mathbf{s}_t$

\subsection{Results}

\begin{figure}
\centering
\includegraphics[width=0.8\textwidth]{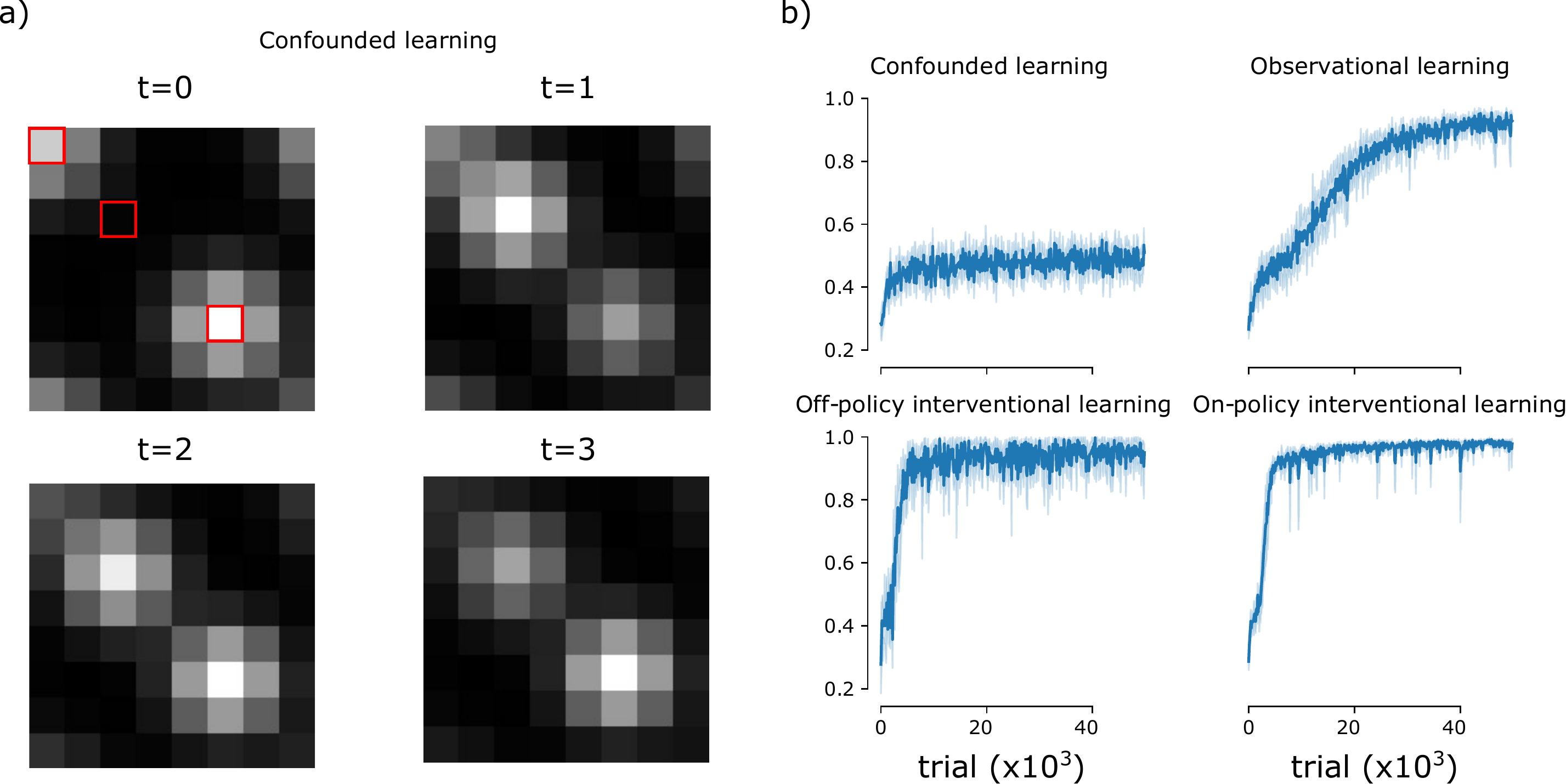}
\caption{Environment with latent causal structure. a) Four frames from the confounded environment condition. Three, blurred, flashing pixels indicate the state of the variables $\mathbf{y}_t$, obeying either chain or delayed fork dynamics between them. For illustration, these are highlighted in red, though this information isn't provided to the agent. The activated pixels decay over time. The other settings are similar, with additional flags to represent the state of $\mathbf{z}_t$, where relevant. b) Curves show mean reward plus/minus standard error over $n = 10$ runs.}
\label{fig:latentresults}
\end{figure}

The meta-RL agent's architecture is modified to add a fully connected layer of size 64, with inputs over the flattened image, before being fed to the LSTM. With this change, the learner is able to quickly learn the correct model. In the interventional conditions, it can learn how to figure out the correct model after only a few thousand trials. The observational setting takes closer to fifty thousand trials to reach the same performance (Figure \ref{fig:latentresults}b). Nonetheless, the agent is able to learn the latent causal structure underlying the observations in all but the confounded case. 

\section{Causal learning in agent-centric action spaces}

We now focus on the setting that relaxes Assumption 2. That is, the action space is now ego-centric, and cannot be interpreted as a direct manipulation of the relevant causal variables underlying the model dynamics. To test causal learning in this setting, we use a simple grid-world environment. 

\subsection{Task details}

The task is an `escape room' task. The environment consists of a 5x5 grid-world. On three edges of the grid are buttons. When anything overlaps with the pixel adjacent to the button, it is pushed. In a given trial of $N$ steps, only one of the buttons (chosen uniformly at random) will activate/open the door on the fourth wall. This door will stay open for $T$ time steps.

There are two phases to this environment. In the first observational phase, for $N_O$ steps, a white object bounces randomly between the three buttons, occasionally activating the door. The agent's actions are ineffective. In the second phase, the action phase, a green cue in the upper left corner indicates that the agent's actions now effect the environment -- they move a separate (gray) object around the grid-world (Figure \ref{fig:egocentric}a). While the bouncing object can move one pixel in the cardinal directions, or one pixel diagonally, the agent can only move one pixel in the four cardinal directions. This phase lasts $N_A$ steps. The agent is rewarded if it moves to the pixel immediately adjacent to the open door. A reward of $R_d$ is administered. 

We test the case that $N_A$ is too small for a valid policy to be the agent, in the action phase, tries each button sequentially to see which one opens the door -- to solve the task it has to pay attention to the observation phase state transitions. Here $N_O = 20, N_A = 10$, the door is open for $T = 5$ time steps, and the reward is $R_d = 10$.

\subsection{Results}

\begin{figure}
\centering
\includegraphics[width=0.6\textwidth]{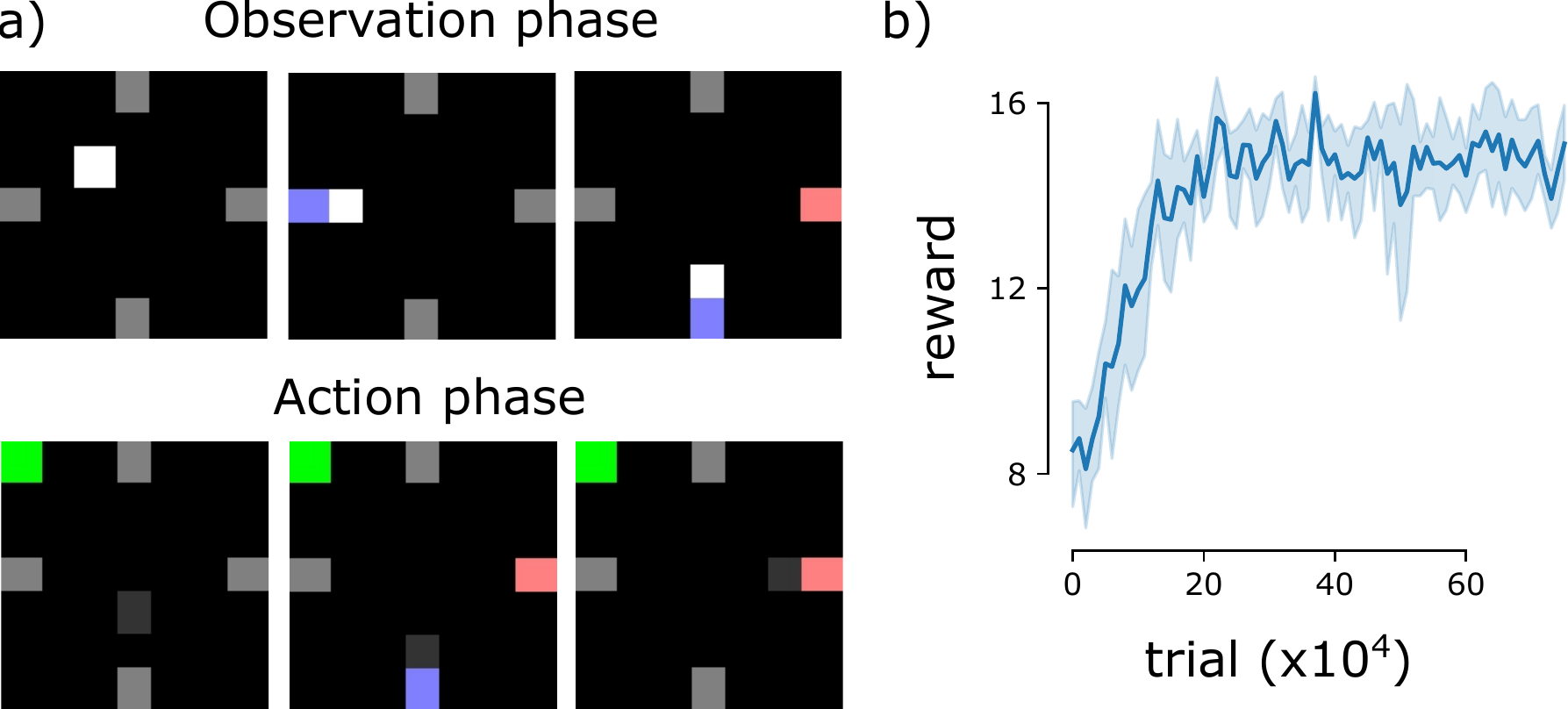}
\caption{The escape-room environment. a) Task involves observation phase, in which the white box bounces randomly between the three buttons (all but the right). When the box is adjacent to the randomly chosen button that opens the door for that trial, it turns red. In the action phase, the agent moves the gray box around. It has to learn from the previous phase which button to go for to open the door and move to it in time to get reward. Some frames are skipped for simplicity in this illustration. b) Curves show mean reward plus/minus standard error over $n = 5$ runs.}
\label{fig:egocentric}
\end{figure}

The same network architecture and algorithm as the previous section is used. With this, the meta-RL agent is able to learn to perform this causal learning task after about two hundred thousand trials (Figure \ref{fig:egocentric}b). The agent start location is randomly chosen, at the start of each action phase. It then successfully moves to the correct button, and then to the door for reward on a high proportion of trials. Thus this setup is able to successfully utilize information from the observation phase for use in the action phase. In doing so the agent is not mimicking the movement of the white box; it is not performing imitation learning. This transfer from observation to action is a key indicator of intervention-centric causal learning.

\section{Discussion}

In essence here we proposed to take a step back from a causal formalism such as Pearl's \citep{pearl2000causality} and ask how behaviors that constitute interventions and observational learning as studied in cognitive science can be recapitulated in a learning agent. Interventions in this sense are a more vague notion than in a statistical causal model, which takes the exact mechanism of the intervention as given. But the human and animal cognitive science literature has studied interventions in many settings and stages of development. So we can better understand what behaviors constitute sophisticated causal reasoning by turning to these studies. 

\subsection{Inspiration from cognitive science}

What notion of causality do we and other animals possess? Philosopher James Woodward \citep{Woodward2010} argues that a notion of interventions is unique to humans. Other animals have either an ego-centric notion of causality -- they are only capable of learning causal relationships as they are revealed by and apply to the agent; or an agent-centric notion of causality, in which learning the structure of the world can also be achieved through reproducing another agent's actions.

Operant conditioning is of course ubiquitous among animals, and thus so is agent-centric causal learning. This type of learning can be quite sophisticated: tool-using animals  demonstrate a nuanced understanding of the consequences of their actions \citep{Schloegl2017}, and model-based reinforcement learning involves learning a causal model of the environment \citep{Gershman2015}. Yet such learning does not require a notion of an environment with structure independent of the agent's actions, and thus no sense in which actions are interventions on that environment.

Social learning amongst animals too can be quite complicated. In imitative learning, the actions of another agent are copied to achieve some goal. In emulative learning, an observed action is not necessarily reproduced, but an action is taken to reproduce a desired, observed outcome \citep{Hopper2010}. However, the causal learning that occurs in these cases may not be perfect: animals may not show an understanding of the relevant parts of another agent's actions to copy, leading to over-imitation -- performing a sub-optimal action simply because the demonstrator did, while control animals that only try the task for themselves learn the optimal action more quickly \citep{Hopper2010}. For these reasons, social learning is also not indicative of a full causal understanding of the environment either. 

These can be contrasted with an intervention-centric notion of causality. The hallmarks of which are the ability to appropriately imitate and emulate others' actions to produce a desired goal, to not over-imitate, to not misuse tools or select the wrong tool for the job, to produce novel interventions that are suggested by relationships learned from observation, and to learn and integrate relationships from a diverse range of sources (not just from others). Though animals such as primates and corvids, and children, can perform some of these \citep{Tennie2010a,Meltzoff2012,Bonawitz2010,Hopper2008,Schloegl2017,Volter2016,Bonawitz2010,Taylor2015,Taylor2014,Jelbert2014,Taylor2007}, adult humans can most robustly perform each of these things.

\subsection{Related work and outlook}

Here we focused on a notion of causal learning relatively unaddressed in the learning agent literature. We provided a simple environment that gets at this issue, and one simple meta-RL algorithm that can solve it. However we do not propose any algorithmic innovations: there may be better approaches, and these may be closely related to those already in use. It's thus useful to highlight related work.

The environment presented here requires the agent to take advantage of data drawn from an observational setting, in which the actions of the agent are ineffective. This is related to learning problems where the action labels generating the observed state sequences are not provided. This setting has been explored recently in imitation learning settings, known as imitation learning from observation (ILO) \citep{Torabi2019, Torabi, Ho2016, Zona2019, Zolna2019, Torabi2018, Li2018, Liu2018, Wu, Edwards2018}. But these methods do not solve the problem by themselves -- they just imitate. Imitating for the sake of it, over-imitation, is a sign of a lack of causal understanding. Recent work combining imitation learning and reinforcement learning, or learning from imperfect demonstrations \citep{Gao2018}, address this to some extent \citep{Zona2019}. RILO does so in a setting where action labels are not provided \citep{Zona2019}. However such approaches, when no reward signal is available, will default to imitation learning. This may be a practical learning approach in the presence of an expert, but it does not get at causal learning. Though some version of these works may prove promising in the task domain we have tested here.

Alternatively we can approach the problem by eschewing imitation learning, and viewing the setting presented here as doing a form of off-policy reinforcement learning where action labels are not provided (similar to \cite{Borsa2017}). Some of the approaches used in ILO may prove still useful. For instance, models based on inverse dynamics models could take the observational data \citep{Christiano2016,Pathak2017}, use the inverse model to infer what action the agent \emph{could} have taken, and then run an RL algorithm using the inferred actions. A sort of vicarious learning, evocative of agency theories of causation \citep{Woodward2010}. A comparison between this approach and that taken here is future work.

A caveat is that here we have just focused on the idea of transferring what is learnt from observation to an action phase. But, in addition to this transferability, a key notion in intervention-based learning is that interventions are specific -- they only act on a particular object. An algorithm's ability to parse the environment into a discrete set of objects, and to learn how these could be manipulated individually, was not tested. This problem relates to recent work on learning environment affordances and controllable factors \citep{Thomas2017,Sawada2018}. Methods like Recurrent Independent Mechanisms \citep{Goyal2019} have proposed learning separable components of an environment with their own autonomous dynamics, which may relate to causal factors in the environment, and thus may prove useful in this regard. 

Closely related to ideas of transferring from observational to action settings is work in the causal inference literature that learns causal models from unknown or uncertain interventions. In invariant causal prediction \citep{Peters2016, Arjovsky}, for instance, robustness to changes in environment are used as a cue for causal relations. These ideas has been explored in a meta-learning setting too \citep{Bengio2019a}. Emerging ideas of the importance of invariance needs to be integrated with this intervention-centric notion used here \citep{Arjovsky,Scholkopf2019}. Finally, and perhaps most related to the work here is work on meta-learning causal learning algorithms \citep{Dasgupta2018}. This work differs from that here in that the environment is low-dimensional, almost fully observed, not dynamic, and the actions are taken as manipulations of the state variables in the causal graph. Thus there is a lot of progress in all of these related areas. A more explicit testing of these learning agents' causal learning abilities in the settings such as those tested here may prove useful in providing AI with a more human-like notion of causality.

\section*{Acknowledgements}

For many useful discussions, the author thanks Anna Leshinskaya and Anirudh Goyal. 

\medskip
\small
\bibliographystyle{plainnat}
\bibliography{Writeups-RDD,Neuro-Observational-learning,Writeups-bandits,Writeups-Learning_causal_inference}

\appendix

\section{Parameters for environments}

\subsection{Simple environment parameters}

Trial lasts $N = 20$ steps. The probability of a given trial having the chain structure is 0.5. Spontaneous rates of activation for $s^i$: $p_1 = 0.1, p_2 = 0.01, p_3 = 0.01$. For the off-policy settings with the `interventional' variables ($z^i$), these are activated with probabilities $p_2^{int} = 0.1, p_3^{int} = 0.1$. Only the second and third variables are perturbed -- a perturbation of the parent node $s^1_t$ will not help discriminate between the two Models. 
\subsection{Visual environment parameters}

The underlying dynamics have the same parameters as the basic environments. 

\subsection{Agent-centric environment parameters}

The parameters of the environment are as follows. Observation and action phase lengths: $N_O = 20, N_A = 10$. Reward administered if successful: $R_d = 10$. Door open for $T = 5$ time steps. 

\end{document}